# Enhancing Diagnostic Accuracy through Multi-Agent Conversations: Using Large Language Models to Mitigate Cognitive Bias

Yu He Ke, MBBS[1,2], Rui Yang, MS[2], Sui An Lie, MMed[1], Taylor Xin Yi Lim, MBBS[1], Hairil Rizal Abdullah, PhD[1,2], Daniel Shu Wei Ting, PhD[2,3,4], Nan Liu, PhD[2,5,6*]

[1] Department of Anesthesiology, Singapore General Hospital, Singapore, Singapore
[2] Center for Quantitative Medicine, Duke-NUS Medical School, Singapore, Singapore
[3] Singapore Eye Research Institute, Singapore National Eye Center, Singapore, Singapore
[4] Byers Eye Institute, Stanford University, Stanford, CA, USA
[5] Program in Health Services and Systems Research, Duke-NUS Medical School, Singapore, Singapore
[6] Institute of Data Science, National University of Singapore, Singapore, Singapore

[*]Correspondence: Nan Liu, Center for Quantitative Medicine, Duke-NUS Medical School, 8 College Rd, Singapore 169857. Phone: +65 6601 6503. Email: liu.nan@duke-nus.edu.sg.

## Abstract

**Background:** Cognitive biases in clinical decision-making significantly contribute to errors in diagnosis and suboptimal patient outcomes. Addressing these biases presents a formidable challenge in the medical field.

**Objective:** This study explores the role of large language models (LLMs) in mitigating these biases through the utilization of a multi-agent framework. We simulate the clinical decision-making processes through multi-agent conversation and evaluate its efficacy in improving diagnostic accuracy.

**Methods:** A total of 16 published and unpublished case reports where cognitive biases have resulted in misdiagnoses were identified from the literature. In the multi-agent framework, we leveraged GPT-4 to facilitate interactions among four simulated agents to replicate clinical team dynamics. Each agent has a distinct role: 1) To make the final diagnosis after considering the discussions, 2) The devil's advocate and correct confirmation and anchoring bias, 3) The tutor and facilitator of the discussion to reduce premature closure bias, and 4) To record and summarize the findings. A total of 80 simulations were evaluated for the accuracy of initial diagnosis, top differential diagnosis and final two differential diagnoses.

**Results:** In a total of 80 responses evaluating both initial and final diagnoses, the initial diagnosis had an accuracy of 0% (0/80), but following multi-agent discussions, the accuracy for the top differential diagnosis increased to 71.3% (57/80), and for the final two differential diagnoses, to 80.0% (64/80).

**Conclusions:** The framework demonstrated an ability to re-evaluate and correct misconceptions, even in scenarios with misleading initial investigations. The LLM-driven multi-agent conversation framework shows promise in enhancing diagnostic accuracy in diagnostically challenging medical scenarios.

## Introduction

Human cognitive biases in clinical decision-making are increasingly recognized as a crucial factor in healthcare errors and suboptimal patient outcomes [1]. These biases stem from innate psychological tendencies and can potentially lead to misjudgments and suboptimal outcomes in patient care [2–4]. Despite concerted efforts involving educational strategies, optimal work environments, and a culture promoting bias awareness and correction [5,6], the eradication of these biases remains an elusive goal.

The integration of artificial intelligence (AI), and in particular, large language models (LLMs), into clinical medicine is on the horizon [7–9]. LLMs have advanced text generation capability and extensive domain-specific knowledge [10]. Notably, these models have demonstrated their proficiency by successfully passing advanced medical examinations [11] and scoring clinical risk gradings on par with experienced physicians [12].

However, the deployment of LLMs in actual clinical diagnosis and decision-making processes has been mired in controversy, primarily due to the high stakes involved. The use of AI in medical settings is not just a technological issue; it intersects with complex ethical, legal, and medical considerations [12–16]. The accuracy of ChatGPT making the correct emergency medicine diagnosis is still limited to 77-83% [17]. Thus, concerns centering around the legal implications and accountability in cases where AI-driven diagnostics might lead to errors or misjudgments are a major hurdle.

This debate is rooted in the fundamental difference between human and machine intelligence. While LLMs can process and analyze vast quantities of data far beyond human capacity, they lack the nuanced understanding, empathy, and ethical reasoning inherent to human practitioners [18]. Human cognitive biases can be mitigated through a combination of awareness, education, and structured approaches [19]. Simulations of such discussions through LLM-agents could provide a new solution to increase the accuracy of diagnosis [20]. Multi-agent framework features dialogue agents with near-human performance and could introduce an innovative paradigm in healthcare [21–23]. By simulating interactive scenarios that mirror real-life clinical decision-making processes, and through reading the multi-agent conversations, clinicians can be made aware of potential cognitive biases and how to correct them. This facilitates learning in a controlled, educational environment [24,25].

Despite the remarkable advancements in LLM technology, particularly in the multi-agent sphere, its application in pinpointing and correcting human cognitive biases in clinical settings is relatively uncharted territory in existing research. This study aims to examine the accuracy of clinical diagnoses made by multi-agent framework after correction for cognitive biases.

## Methods

We accessed GPT-4 [26] through an action programming interface (API) call to the OpenAI server. The specific variant utilized was GPT-4 Turbo. Due to the nature of the study, institutional review board approval was not required, as the research did not involve patient data and did not constitute human subjects research.

We implemented a comprehensive search strategy aimed at including all relevant reports on misdiagnoses attributed to cognitive biases. A selection of 15 case reports was identified after a full review of the published literature as representative sampling.

### Search Strategy

In this study, we focused on case reports highlighting instances of misdiagnoses resulting from cognitive biases. Our research involved a comprehensive search of the PubMed database using the terms "case reports[Publication Type]" AND "cognitive bias[All Fields]". Eligibility for inclusion requires case reports to meet four key criteria: 1) They must provide detailed case information sufficient for making the initial diagnosis; 2) They must include a final, accurate diagnosis for the patient; 3) The incorrect diagnosis must be linked to cognitive bias by the authors; and 4) The final diagnosis should not be a rare disease or unclear. A rare disease is a disease or condition that affects fewer than 200,000 patients per year. The list of exclusions for rare diseases was based on the National Organization for Rare Disorders [27]. We set no limits on the publication year of these reports.

Screening of abstracts for eligible studies was conducted by two independent clinically trained reviewers, KYH and TLXY. Each reviewer separately assessed whether a case should be included or excluded based on predefined criteria. In instances of disagreement, SAL reviewed the justifications for exclusion and inclusion and made the final decision. The full texts were reviewed to obtain the case summary, the initial wrong diagnosis by the medical team, and the final diagnosis of the case reported.

For the studies included in the analysis, full-text extraction including patient demographics, past medical history, initial presenting complaints, and results of the preliminary investigations was conducted. For cases involving imaging data, such as Chest X-rays, we did not incorporate the actual images into the query. Instead, we opted to include the legends or descriptions accompanying these images. In defining the boundaries of the clinical scenarios for our study, we restricted the information to that available up to the point of and before the initial diagnosis. This meant deliberately excluding any subsequent investigations, treatments, or management strategies that followed.

As GPT-4 Turbo has a knowledge base trained up to April 2023 [28], there is potential bias stemming from the inclusion of case reports that might have been part of the LLM's pre-training data. Hence a personal clinical scenario that was not published on the internet was included. This complex case was derived from the critical care attending's personal experience where cognitive biases had resulted in wrong and delayed diagnosis. A concise summary of these clinical scenarios, including the unique case, is provided in Multimedia Appendix 1.

### Multi-Agent Conversation Framework

In this study, we utilized the multi-agent conversation framework provided by AutoGen [21] to assess its efficacy in mitigating cognitive biases in clinical decision-making. Within the system, each agent interacts based on their predefined role prompts, thereby simulating the collaborative decision-making process typically observed among healthcare professionals.

The suggested optimal group size to facilitate group discussion and performance has been proposed to be between 3–5 [29]. In the absence of established literature recommending an optimal team size for mitigating cognitive biases in medical settings, we constructed a simulation using four agents, representing a typical clinical team composition [30], as depicted in (Figure 1). This configuration aims to realistically emulate the dynamics of clinical teams and their potential to reduce cognitive biases.

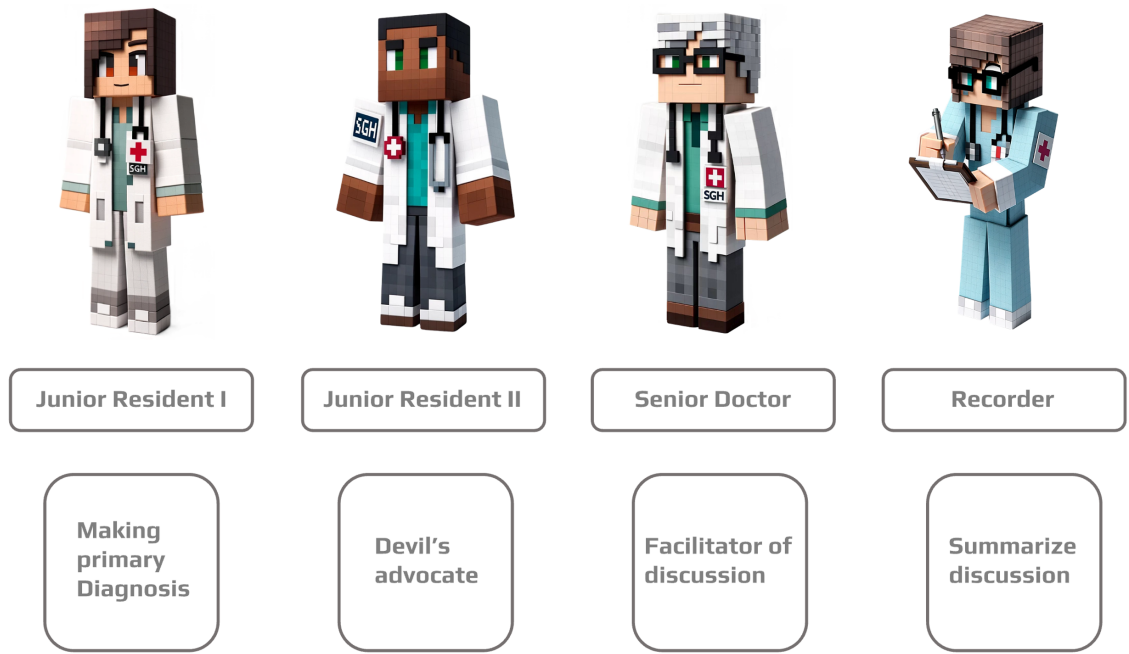

Figure 1. Different roles in the multi-agent conversation framework.

The diagnostic process was orchestrated through the collaborative efforts of simulated medical professionals (agents) with varying levels of expertise. Junior Resident I (JR1), as the primary physician, was tasked with presenting the initial diagnosis. JR1 was given the personality of making swift assumptions but is willing to embrace feedback and consider alternative diagnostic possibilities. After the group discussion, JR1 is then allowed to re-consider the most probable differential diagnosis along with an alternative. Junior Resident II (JR2), a colleague of JR1, critically appraised the initial diagnosis, pinpointing inconsistencies and advocating for alternative differential diagnoses. This role was instrumental in addressing potential confirmation and anchoring biases in the diagnostic process. Complementing the juniors, the Senior Doctor (SD) brought in-depth experience to the table, crucially identifying cognitive biases in the initial diagnosis and steering the junior doctors toward a more nuanced and accurate diagnosis. This guidance was vital in circumventing premature diagnostic closure and knowledge bias. Additionally, the role of the Recorder was to distill the outcomes of the discussion, compiling a definitive list of differential diagnoses and extracting key learning points from the collaborative effort, thereby enriching the diagnostic process with a comprehensive and multifaceted approach.

### Diagnostic Accuracy Assessment

The overall performance of the framework was evaluated based on the accuracy of 1) the first most likely differential diagnosis and 2) the final two differential diagnoses. The diagnostic accuracy of the multi-agent system was evaluated by comparing the final diagnosis made by the agents to the actual patient outcomes, as established through subsequent patient management and results. For this purpose, each diagnosis was categorized as either "Correct" or "Incorrect". In addition, to assess the consistency and reliability of the agents, each clinical scenario was simulated five times. This repetitive analysis allowed for the extraction and examination of variations in the agents' responses, offering insights into the consistency of the diagnoses across multiple iterations.

### Bias Identification and Mitigation

An integral part of the evaluation involved documenting the specific cognitive biases identified and addressed during the agents' discussions. This aspect focused on understanding how effectively the multi-agent system could recognize and mitigate cognitive biases, which are crucial factors in diagnostic accuracy. As cognitive biases are subjective and many biases may be present within the same scenario, the identification of cognitive bias within the diagnosis was evaluated for hallucinations and is negatively penalized. If, within the discussions, there are inappropriate cognitive bias discussions present, the scenario will be highlighted and marked as inaccurate for the presence of hallucinations. The interaction and decision-making process among the agents are illustrated in (Figure 2). This representation aids in visualizing the dynamics of the simulation and the interplay between different agents in reaching a diagnosis.

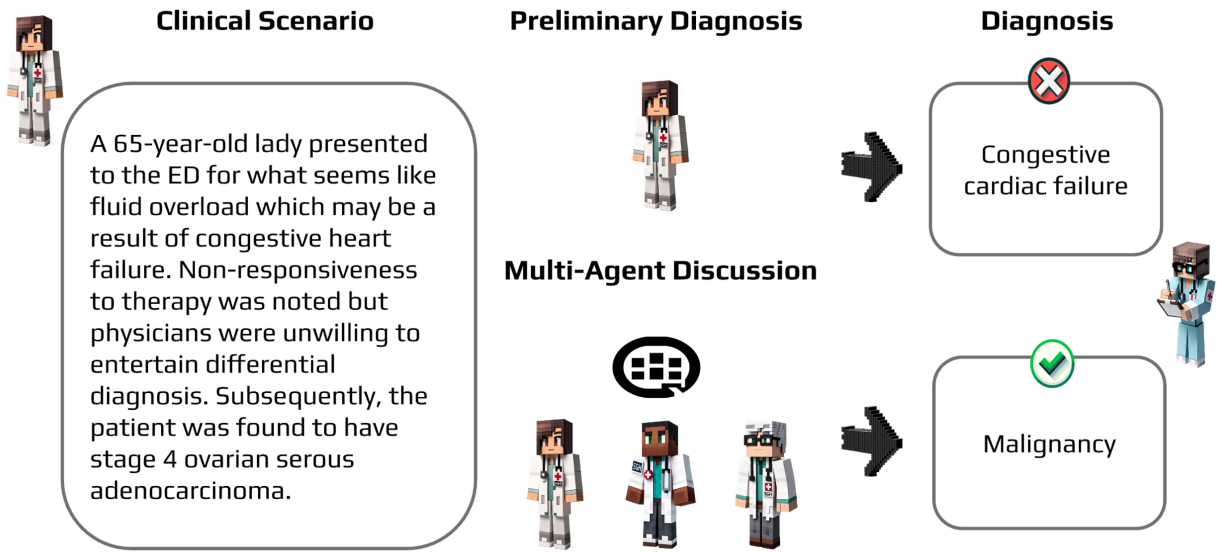

Figure 2. Schematic illustration of multi-agent discussion dynamics leading to accurate differential diagnosis.

## Results

A comprehensive search of the PubMed database yielded a total of 162 case reports, of which 37 of these were determined to be eligible for inclusion in the study. From this subset, 15 cases were selected for evaluation as representative sampling. Additionally, a 16th scenario, derived from a critical care attending personal clinical experience was included. The PRISMA (Preferred Reporting Items for Systematic Review and Meta-Analysis) diagram can be viewed in Multimedia Appendix 2.

### Overall Performance of Multi-Agent Conversation Framework

A total of 80 responses were generated by the multi-agent conversation framework, encompassing both initial and final diagnoses following discussions with the agents. The initial diagnosis made by the first-responder agent, JR1, had a 0% correctness rate compared to the final correct diagnosis established by the clinical scenario. However, after multi-agent discussions, the accuracy of the top differential had a 57/80 (71.3%) correct rate, while the two differentials had a correct rate of 80.0% (64/80).

### Clinical Scenarios

The clinical cases covered a broad spectrum of medical fields, ranging from pediatric to malignancy diagnosis. Specifically, 6 cases were centered on infectious disease diagnosis, 3 pertained to critical care, and 2 involved vascular-related diagnoses. The rest were diverse, spanning neurology, gynecology, cancer, urology, and endocrinology. A notable aspect of these cases was the cross-disciplinary nature of most initial and final diagnoses, observed in 12 out of the 16 cases. For example, in one illustrative case (case 1 [31]), an elderly patient presented with symptoms of shortness of breath on exertion and cough, leading to an initial misdiagnosis of heart failure. However, further investigation, considering her ongoing treatment with infliximab for rheumatoid arthritis, revealed the actual diagnosis of miliary tuberculosis. Another case (case 12 [32]) involved a young woman presenting with sudden, left-sided sharp pleuritic chest pain, which lessened when sitting forward. Despite the initial chest radiograph being interpreted as showing no acute abnormalities, the AutoGen system, provided with this potentially misleading information, initially diagnosed the condition as a pulmonary embolism. Yet, after a thorough discussion and re-evaluation, the correct diagnosis of pneumothorax was established, indicating a missed finding in the chest radiograph.

### Consistency of Multi-Agent Conversation Framework

There were variations observed in the repeated scenarios, particularly in the process of generating the most likely diagnosis (Kappa Coefficient = 0.838) versus the top two differential diagnoses (0.913), although the overall agreement for all diagnoses given was high. For instance, in case 13 [33], a young patient presenting with lactic acidosis was initially diagnosed with diabetic ketoacidosis and further discussions within the multi-agent environment led to the identification of thiamine deficiency. In 3 out of 5 simulations (60%), JR1 identified thiamine deficiency as the top differential diagnosis following the multi-agent discussions. In the remaining two simulations, gastrointestinal disorders were initially considered the most likely diagnosis, with thiamine deficiency being the second most likely differential.

The multi-agent discussions have helped JR1 to reach the final correct diagnosis in 13 out of the 16 scenarios. Furthermore, the discussion was able to identify various clinical biases present - such as anchoring bias, confirmation bias, availability bias, and premature closure. A detailed breakdown of the diagnostic performance is presented in Table 1, with the detailed breakdown of the answers and biases identified given in Multimedia Appendix 1.

Table 1. Responses from the multi-agent conversation framework, the initial impression was given by Junior Resident I, and the corrected diagnosis was recorded after discussion by various agents.

| | Publication Year | Initial Wrong diagnosis | Final Correct Diagnosis | Initial diagnosis correct | Top differential correct | Final differential correct | Hallucinations of bias identified |
|---|---|---|---|---|---|---|---|
| 1 [31] | 2015 | Heart failure | Miliary tuberculosis | 0/5 (0%) | 0/5 (0%) | 0/5 (0%) | None |
| 2 [34] | 2017 | Bronchial asthma triggered by bacterial pneumonia | Heart failure secondary to dilated cardiomyopathy | 0/5 (0%) | 0/5 (0%) | 0/5 (0%) | None |
| 3 [35] | 2018 | Syndrome of inappropriate secretion of Antidiuretic hormone (SIADH) | Non-functioning macro pituitary adenoma causing adrenal insufficiency | 0/5 (0%) | 2/5 (40%) | 5/5 (100%) | None |
| 4 [36] | 2019 | Headache and neck pain - migraines and muscle strain | cryptococcal meningitis | 0/5 (0%) | 5/5 (100%) | 5/5 (100%) | None |
| 5 [37] | 2020 | Complex regional pain syndrome flare | Left common and external iliac artery occlusion | 0/5 (0%) | 5/5 (100%) | 5/5 (100%) | None |
| 6 [38] | 2022 | pelvic inflammatory disease | Atypical ectopic pregnancy | 0/5 (0%) | 5/5 (100%) | 5/5 (100%) | None |
| 7 [39] | 2022 | COVID-19 pneumoniae | Bacterial pneumonia (Legionella pneumoniae) | 0/5 (0%) | 5/5 (100%) | 5/5 (100%) | None |
| 8 [40] | 2022 | Urinary tract infection with complicated pyelonephritis | Vesicointestinal fistula due to Crohn's disease | 0/5 (0%) | 5/5 (100%) | 5/5 (100%) | None |
| 9 [41] | 2022 | Bone (sternum) Tuberculosis | Syphilitic gumma and osteomyelitis | 0/5 (0%) | 5/5 (100%) | 5/5 (100%) | None |
| 10 [42] | 2022 | Urinary tract infection | Vertebral osteomyelitis and bilateral psoas and retroperitoneal abscesses | 0/5 (0%) | 3/5 (60%) | 5/5 (100%) | None |
| 11 [43] | 2022 | Anaphylaxis secondary to Henna | Superior vena cava syndrome | 0/5 (0%) | 5/5 (100%) | 5/5 (100%) | None |
| 12 [32] | 2023 | Pulmonary embolism | Pneumothorax | 0/5 (0%) | 3/5 (60%) | 5/5 (100%) | None |
| 13 [33] | 2023 | Diabetic ketoacidosis with infections | thiamine deficiency | 0/5 (0%) | 3/5 (60%) | 5/5 (100%) | None |
| 14 [44] | 2023 | Endometritis | Ischemic bowel | 0/5 (0%) | 0/5 (0%) | 0/5 (0%) | None |

| 15 [45] | 2023 | Acute myocarditis likely due to MIS-C | Acute myocarditis caused by invasive bacterial infection | 0/5 (0%) | 1/5 (20%) | 4/5 (80%) | None |
| 16 | - | Congestive cardiac failure | Malignancy | 0/5 (0%) | 5/5 (100%) | 5/5 (100%) | None |
| Total | - | | | 0/80 (0%) | 57/80 (71.3%) | 64/80 (80.0%) | None |

## Discussion

### Principal Results

This study assessed the effectiveness of the multi-agent conversation framework in improving diagnostic accuracy and mitigating cognitive biases in clinical decision-making. Our findings reveal that the integration of multi-agent discussions substantially enhances diagnostic accuracy with the top differential diagnosis correctness rate reaching 71.3%, and the final two differential diagnoses achieving an 80.0% correctness rate, underlining the potential of AI in healthcare.

The increase in diagnostic accuracy demonstrates the value of multi-perspective analysis in medical diagnosis, a core feature of the multi-agent conversation environment. This is in line with previous research emphasizing the importance of collaborative decision-making in healthcare to mitigate individual cognitive biases [46]. The consistency of responses in the repeated scenarios further validates its reliability and potential applicability in real-world clinical settings.

The effectiveness of the multi-agent conversation system, particularly in scenarios involving misleading or misinterpreted initial investigations, is noteworthy. This was exemplified in case 12, where our multi-agent framework successfully identified a pneumothorax that had been initially overlooked by human clinicians. The LLM-agent was able to critically examine and question potential misinterpretations. Such capabilities are crucial in refining the diagnostic process and enhancing accuracy. While Brown et al. discussed the role of artificial intelligence (AI) in the identification of pneumothorax [32], there is a potential for pre-trained LLMs to replace the decision aid, rather than to develop new resource-intensive systems such as image deep learning. This strategy aligns with the current trajectory of AI development in healthcare, where the focus is on integrating and maximizing existing AI technologies to enhance clinical decision-making and focus on sustainable AI [47].

Furthermore, the inclusion of a wide range of medical fields and cross-disciplinary diagnoses in our selected cases provided a robust test of the multi-agent system's capabilities. The system's performance across varied clinical scenarios suggests its potential applicability in diverse medical contexts. The ability to reach the correct diagnosis when tasked to only give 2 differential diagnoses is a challenging task, in which the multi-agent environment excelled. This versatility is crucial for any AI system intended for use in real-world clinical environments, which are inherently multidisciplinary and complex.

The results of our study extend beyond the educational benefits of multi-agent, highlighting their potential for broader clinical integration. The reflective process fostered by engaging with LLMs in diagnosing and revising cases not only cultivates an educational atmosphere conducive to developing critical thinking skills but also suggests practical applications in clinical settings [48]. One notable avenue is the integration of multi-agent into Electronic Medical Records systems. This could enhance decision-making processes by providing real-time, data-driven insights and augmenting the cognitive capabilities of medical professionals. Such integration would not only streamline the diagnostic process but also aid in the identification of potential cognitive biases, thereby enhancing the quality of patient care. Furthermore, the incorporation of multi-agent in EMRs could facilitate continuous learning and improvement, ensuring that medical practitioners remain updated with the latest medical knowledge and best practices, crucial for maintaining high standards in patient treatment and care.

## Limitations

The study, while providing valuable insights into the potential application of multi-agent in clinical diagnostics, is subject to several limitations. Firstly, the reliance on published case reports limits the breadth of clinical scenarios, potentially affecting the generalizability of our findings to broader medical practice. Secondly, the exclusion of visual data, such as medical imaging, confines our model's diagnostic capabilities to text-based information, omitting a critical component of clinical diagnosis. Additionally, the inherent biases present in the LLMs, based on its pre-training data, could have influenced the diagnostic suggestions. Meanwhile, the technical limitations inherent in LLMs, including their understanding of complex medical terminologies and nuances, may not match the expertise of experienced clinicians, possibly limiting the scope of applicability.

Future studies could assess the effectiveness and adaptability of the multi-agent framework in evolving clinical scenarios. More importantly, while LLMs have demonstrated potential as a valuable clinical aid in correcting cognitive biases, the implementation of such technology in healthcare necessitates rigorous ethical and regulatory oversight [49] and should continue to augment rather than replace the human clinician's expertise [15].

## Conclusions

In conclusion, our study highlights the potential of multi-agent in enhancing diagnostic accuracy in clinical decision-making. The findings support the integration of advanced generative AI technology in educational and clinical settings as a tool for augmenting human decision-making. Future research should focus on the application of these systems in real-time clinical environments and their impact on patient outcomes. Moreover, ethical and legal considerations regarding the use of AI in healthcare need continued exploration to ensure patient safety and professional accountability.

## Acknowledgements
This study was supported by the Duke-NUS Signature Research Program funded by the Ministry of Health, Singapore. Any opinions, findings and conclusions or recommendations expressed in this material are those of the author(s) and do not reflect the views of the Ministry of Health.

## Authors' Contribution
YHK and SAL conceived the study. YHK, RY, DSWT and NL designed the study. YHK and RY drafted the manuscript with critical appraisal and further development by DSWT, NL YHK, TXYL and RY conducted data analyses. NL supervised the study. All authors contributed to the revision of the manuscript and approval of the final version.

## Conflicts of Interest
None declared.

## Abbreviations
AI: artificial intelligence
API: action programming interface
LLMs: large language models
PRISMA: preferred reporting items for systematic review and meta-analysis

# References

1. Saposnik G, Redelmeier D, Ruff CC, Tobler PN. Cognitive biases associated with medical decisions: a systematic review. BMC Med Inform Decis Mak 2016 Nov 3;16(1):138. PMID:27809908
2. Korteling JEH, Paradies GL, Sassen-van Meer JP. Cognitive bias and how to improve sustainable decision making. Front Psychol 2023 Feb 28;14:1129835. PMID:37026083
3. Berthet V. The Impact of Cognitive Biases on Professionals' Decision-Making: A Review of Four Occupational Areas. Front Psychol 2021;12:802439. PMID:35058862
4. Beldhuis IE, Marapin RS, Jiang YY, Simões de Souza NF, Georgiou A, Kaufmann T, Castela Forte J, van der Horst ICC. Cognitive biases, environmental, patient and personal factors associated with critical care decision making: A scoping review. J Crit Care 2021 Aug;64:144–153. PMID:33906103
5. Doherty TS, Carroll AE. Believing in Overcoming Cognitive Biases. AMA J Ethics 2020 Sep 1;22(9):E773–778. PMID:33009773
6. Hershberger PJ, Markert RJ, Part HM, Cohen SM, Finger WW. Understanding and addressing cognitive bias in medical education. Adv Health Sci Educ Theory Pract 1996 Jan;1(3):221–226. PMID:24179022
7. Yang R, Tan TF, Lu W, Thirunavukarasu AJ, Ting DSW, Liu N. Large language models in health care: Development, applications, and challenges. Health Care Science John Wiley & Sons, Ltd; 2023 Aug 1;2(4):255–263.
8. Yang R, Zeng Q, You K, Qiao Y, Huang L, Hsieh C-C, Rosand B, Goldwasser J, Dave AD, Keenan TDL, Chew EY, Radev D, Lu Z, Xu H, Chen Q, Li I. Ascle: A Python Natural Language Processing Toolkit for Medical Text Generation. 2023. Available from: http://arxiv.org/abs/2311.16588 [accessed Jan 17, 2024]
9. Yang R, Liu H, Marrese-Taylor E, Zeng Q, Ke YH, Li W, Cheng L, Chen Q, Caverlee J, Matsuo Y, Li I. KG-Rank: Enhancing Large Language Models for Medical QA with Knowledge Graphs and Ranking Techniques. 2024. Available from: http://arxiv.org/abs/2403.05881 [accessed Apr 11, 2024]
10. Nori H, Lee YT, Zhang S, Carignan D, Edgar R, Fusi N, King N, Larson J, Li Y, Liu W, Luo R, McKinney SM, Ness RO, Poon H, Qin T, Usuyama N, White C, Horvitz E. Can Generalist Foundation Models Outcompete Special-Purpose Tuning? Case Study in Medicine. 2023. Available from: http://arxiv.org/abs/2311.16452 [accessed Jan 17, 2024]
11. Kung TH, Cheatham M, Medenilla A, Sillos C, De Leon L, Elepaño C, Madriaga M, Aggabao R, Diaz-Candido G, Maningo J, Tseng V. Performance of ChatGPT on USMLE: Potential for AI-assisted medical education using large language models. PLOS Digit Health 2023 Feb;2(2):e0000198. PMID:36812645
12. Lim DYZ, Ke YH, Sng GGR, Tung JYM, Chai JX, Abdullah HR. Large language models in anaesthesiology: use of ChatGPT for American Society of Anesthesiologists physical status classification. Br J Anaesth 2023 Jul 18;

PMID:37474421
13. Wang H, Fu T, Du Y, Gao W, Huang K, Liu Z, Chandak P, Liu S, Van Katwyk P, Deac A, Anandkumar A, Bergen K, Gomes CP, Ho S, Kohli P, Lasenby J, Leskovec J, Liu T-Y, Manrai A, Marks D, Ramsundar B, Song L, Sun J, Tang J, Veličković P, Welling M, Zhang L, Coley CW, Bengio Y, Zitnik M. Scientific discovery in the age of artificial intelligence. Nature 2023 Aug;620(7972):47–60. PMID:37532811
14. Karabacak M, Margetis K. Embracing Large Language Models for Medical Applications: Opportunities and Challenges. Cureus 2023 May;15(5):e39305. PMID:37378099
15. Singhal K, Azizi S, Tu T, Mahdavi SS, Wei J, Chung HW, Scales N, Tanwani A, Cole-Lewis H, Pfohl S, Payne P, Seneviratne M, Gamble P, Kelly C, Babiker A, Schärli N, Chowdhery A, Mansfield P, Demner-Fushman D, Agüera Y Arcas B, Webster D, Corrado GS, Matias Y, Chou K, Gottweis J, Tomasev N, Liu Y, Rajkomar A, Barral J, Semturs C, Karthikesalingam A, Natarajan V. Large language models encode clinical knowledge. Nature 2023 Aug;620(7972):172–180. PMID:37438534
16. Harrer S. Attention is not all you need: the complicated case of ethically using large language models in healthcare and medicine. EBioMedicine 2023 Apr;90:104512. PMID:36924620
17. Berg HT, van Bakel B, van de Wouw L, Jie KE, Schipper A, Jansen H, O'Connor RD, van Ginneken B, Kurstjens S. ChatGPT and Generating a Differential Diagnosis Early in an Emergency Department Presentation. Ann Emerg Med 2023 Sep 9; PMID:37690022
18. Clusmann J, Kolbinger FR, Muti HS, Carrero ZI, Eckardt J-N, Laleh NG, Löffler CML, Schwarzkopf S-C, Unger M, Veldhuizen GP, Wagner SJ, Kather JN. The future landscape of large language models in medicine. Commun Med 2023 Oct 10;3(1):141. PMID:37816837
19. Satya-Murti S, Lockhart J. Recognizing and reducing cognitive bias in clinical and forensic neurology. Neurol Clin Pract 2015 Oct;5(5):389–396. PMID:29443168
20. Nascimento N, Alencar P, Cowan D. Self-Adaptive Large Language Model (LLM)-Based Multiagent Systems. arXiv [csMA]. 2023. Available from: http://arxiv.org/abs/2307.06187
21. Wu Q, Bansal G, Zhang J, Wu Y, Li B, Zhu E, Jiang L, Zhang X, Zhang S, Liu J, Awadallah AH, White RW, Burger D, Wang C. AutoGen: Enabling Next-Gen LLM Applications via Multi-Agent Conversation. arXiv [csAI]. 2023. Available from: http://arxiv.org/abs/2308.08155
22. Wang Z, Zhang G, Yang K, Shi N, Zhou W, Hao S, Xiong G, Li Y, Sim MY, Chen X, Zhu Q, Yang Z, Nik A, Liu Q, Lin C, Wang S, Liu R, Chen W, Xu K, Liu D, Guo Y, Fu J. Interactive Natural Language Processing. arXiv [csCL]. 2023. Available from: http://arxiv.org/abs/2305.13246
23. Shanahan M, McDonell K, Reynolds L. Role-Play with Large Language Models. arXiv [csCL]. 2023. Available from: http://arxiv.org/abs/2305.16367
24. Légaré F, Adekpedjou R, Stacey D, Turcotte S, Kryworuchko J, Graham ID,

Lyddiatt A, Politi MC, Thomson R, Elwyn G, Donner-Banzhoff N. Interventions for increasing the use of shared decision making by healthcare professionals. Cochrane Database Syst Rev 2018 Jul 19;7(7):CD006732. PMID:30025154
25. Vela MB, Erondu AI, Smith NA, Peek ME, Woodruff JN, Chin MH. Eliminating Explicit and Implicit Biases in Health Care: Evidence and Research Needs. Annu Rev Public Health 2022 Apr 5;43:477–501. PMID:35020445
26. ChatGPT. Available from: https://chat.openai.com/ [accessed Sep 16, 2023]
27. Alexander. National organization for rare disorders. National Organization for Rare Disorders. 2022. Available from: https://rarediseases.org/ [accessed Nov 16, 2023]
28. GPT-4. Available from: https://openai.com/research/gpt-4 [accessed Jan 1, 2024]
29. Hackman JR, Vidmar N. Effects of size and task type on group performance and member reactions. Sociometry 1970 Mar;33(1):37–54.
30. Rosen MA, DiazGranados D, Dietz AS, Benishek LE, Thompson D, Pronovost PJ, Weaver SJ. Teamwork in healthcare: Key discoveries enabling safer, high-quality care. Am Psychol 2018 May-Jun;73(4):433–450. PMID:29792459
31. Mull N, Reilly JB, Myers JS. An elderly woman with "heart failure": Cognitive biases and diagnostic error. CCJM Cleveland Clinic Journal of Medicine; 2015 Nov 1;82(11):745–753. PMID:26540325
32. Brown C, Nazeer R, Gibbs A, Le Page P, Mitchell AR. Breaking Bias: The Role of Artificial Intelligence in Improving Clinical Decision-Making. Cureus 2023 Mar;15(3):e36415. PMID:37090406
33. Chehayeb RJ, Ilagan-Ying YC, Sankey C. Addressing Cognitive Biases in Interpreting an Elevated Lactate in a Patient with Type 1 Diabetes and Thiamine Deficiency. J Gen Intern Med 2023 May;38(6):1547–1551. PMID:36814053
34. Tetsuhara K, Tsuji S, Nakano K, Kubota M. Case Report: Heart failure in dilated cardiomyopathy mimicking asthma triggered by pneumonia. BMJ Case Rep BMJ Publishing Group; 2017;2017. PMID:29127129
35. Ilaiwy A, Thompson NE, Waheed AA. Case Report: Adenoma mimicking hyponatremia of SIAD. BMJ Case Rep BMJ Publishing Group; 2018;2018. PMID:30269093
36. Deming M, Mark A, Nyemba V, Heil EL, Palmeiro RM, Schmalzle SA. Cognitive biases and knowledge deficits leading to delayed recognition of cryptococcal meningitis. IDCases Elsevier; 2019;18. PMID:31360635
37. Khawaja H, Font C. Common and external iliac artery occlusion in Behçet's disease: a case of anchoring bias. BMJ Case Rep BMJ Publishing Group; 2020;13(12). PMID:33298479
38. Birch EM, Torres Molina M, Oliver JJ. Not Like the Textbook: An Atypical Case of Ectopic Pregnancy. Cureus 2022 Oct;14(10):e29881. PMID:36348920
39. Kyere K, Aremu TO, Ajibola OA. Availability Bias and the COVID-19 Pandemic: A Case Study of Legionella Pneumonia. Cureus 2022 Jun;14(6):e25846. PMID:35832749
40. Miyagami T, Nakayama I, Naito T. What Causes Diagnostic Errors? Referred

Patients and Our Own Cognitive Biases: A Case Report. Am J Case Rep 2022 Mar 18;23:e935163. PMID:35301273
41. Kamegai K, Yokoyama S, Takakura S, Takayama Y, Shiiki S, Koyama H, Narita M. Syphilitic osteomyelitis in a patient with HIV and cognitive biases in clinical reasoning: A case report. Medicine 2022 Oct 7;101(40):e30733. PMID:36221388
42. Kawahigashi T, Harada Y, Watari T, Harada T, Miyagami T, Shikino K, Inada H. Missed Opportunities for Diagnosing Vertebral Osteomyelitis Caused by Influential Cognitive Biases. Am J Case Rep 2022 Jun 22;23:e936058. PMID:35729859
43. Salama ME, Ukwade P, Khan AR, Qayyum H. Facial Swelling Mimicking Anaphylaxis: A Case of Superior Vena Cava Syndrome in the Emergency Department. Cureus 2022 Sep;14(9):e29678. PMID:36320962
44. Vittorelli J, Cacchillo J, McCool M, McCague A. Cognitive Bias in the Management of a Critically Ill 29-Year-Old Patient. Cureus 2023 May;15(5):e39314. PMID:37351237
45. Stanzelova A, Debray A, Allali S, Belhadjer Z, Taha M-K, Cohen JF, Toubiana J. Severe Bacterial Infection Initially Misdiagnosed as MIS-C: Caution Needed. Pediatr Infect Dis J 2023 Jun 1;42(6):e201–e203. PMID:36916866
46. Lieder F, Griffiths TL, M Huys QJ, Goodman ND. The anchoring bias reflects rational use of cognitive resources. Psychon Bull Rev 2018 Feb;25(1):322–349. PMID:28484952
47. Richie C. Environmentally sustainable development and use of artificial intelligence in health care. Bioethics 2022 Jun;36(5):547–555. PMID:35290675
48. Papathanasiou IV, Kleisiaris CF, Fradelos EC, Kakou K, Kourkouta L. Critical thinking: the development of an essential skill for nursing students. Acta Inform Med 2014 Aug;22(4):283–286. PMID:25395733
49. Meskó B, Topol EJ. The imperative for regulatory oversight of large language models (or generative AI) in healthcare. NPJ Digit Med 2023 Jul 6;6(1):120. PMID:37414860

**Multimedia Appendix 1.** Overview of clinical scenarios and identified cognitive biases as acknowledged in published case reports and by senior physicians.

| Scenario | Summary | Initial diagnosis by clinicians | Final correct diagnosis |
|---|---|---|---|
| 1 | Elderly patient presented with shortness of breath on exertion and cough. She was initially misdiagnosed as having heart failure. Medical reconciliation later reveals that she has been taking infliximab for her RA. Subsequent investigations such as a CT thorax, sputum cultures from a bronchoscopy, and a normal echocardiogram reveal that she has miliary TB. | Heart failure | Miliary tuberculosis |
| 2 | 23 month old girl presenting with wheezing, fever, cough, and rhinorrhoea was initially misdiagnosed as having an asthma attack triggered by pneumonia. The patient's symptoms initially seemed to improve with salbutamol and IV steroid administration, however, the patient's symptoms recurred and she was re-presented to ED multiple times. During the repeat ED visits, the patient was repeatedly diagnosed and treated for asthma due to anchoring bias. Subsequent investigations showed that the patient had heart failure due to dilated cardiomyopathy. | Bronchial asthma triggered by bacterial pneumonia | Heart failure secondary to dilated cardiomyopathy |
| 3 | A 50-year-old male with hyponatremia was wrongly diagnosed with SIADH due to anchoring bias and urine studies that supported their preliminary diagnosis. SIADH should be a diagnosis of exclusion after ruling out other causes such as adrenal and thyroid aetiologies, including hypothyroidism and hypocortisoism. The patient was eventually diagnosed with a pituitary adenoma. | SIADH | Pituitary adenoma |
| 4 | A 33-year-old male with cryptococcal meningitis was misdiagnosed with migraines and muscle strain due to delayed HIV diagnosis failing to recognize immunosuppression. Even with the HIV diagnosis, the team attributed his persistent | Headache and neck pain - migraines | Cryptococcal meningitis |
| | symptoms to muscle strain and migraine due to anchoring bias as he previously responded to symptomatic treatment. | and muscle strain Nausea - gastritis | |
| 5 | 45 year old female with a background of Behcet's disease and complex regional pain syndrome presents with a complex left lower limb hyperalgesia, oedema, and skin temperature changes. She was wrongly diagnosed due to anchoring bias by her physicians as her symptoms seem to fit her already established diagnosis of complex regional pain syndrome. Her team was late to consider an arterial thrombosis although it also fit the patient's symptoms as it is rare. | Complex regional pain syndrome flare | Left common and external iliac artery occlusion |
| 6 | a 25-year-old patient who had an atypical ectopic pregnancy but was misdiagnosed as a pelvic inflammatory disease due to resolution of pain and tachycardia with treatment. | Pelvic inflammatory disease | Atypical ectopic pregnancy |
| 7 | A patient who presented with pneumonia symptoms during the COVID-19 pandemic. He was initially suspected of having COVID. Due to persistent hyponatremia and negative COVID swabs, other differential diagnosis was considered later and tested positive for legionella subsequently. | COVID-19 pneumoniae. | Bacterial pneumonia (Legionella pneumoniae) |
| 8 | 21 year old male was referred from another hospital for what seemed like a poorly responding complicated UTI despite prolonged antibiotic treatment. The physicians failed to investigate the cause of the complicated UTI which turned out to be a vesicointestinal fistula due to Crohn's disease. | Urinary tract infection (UTI) with complicated pyelonephritis | Vesicointestinal fistula due to Crohn's disease |
| 9 | A 30-year-old bisexual male, with a significant history of recurrent syphilis infections, presenting with chest pain and MRI findings of erosive lesions on the sternum and left ninth rib was wrongly treated for bone TB as the physicians failed to consider syphilis osteomyelitis as the patient did not present with any signs of early syphilis. This atypical | Bone (sternum) TB | Syphilitic gumma and osteomyelitis |
| | presentation of syphylis osteomyelitis is typical in HIV patients. | | |
| 10 | 83 year old male presenting with acute on chronic lower back pain and fever was wrongly diagnosed with UTI due to anchoring bias with a pyuria urine sample despite him having no UTI symptoms and multiple lower back pain "red flags". Patient was eventually diagnosed with vertebral osteomyelitis. | Urinary tract infection | Vertebral osteomyelitis and bilateral psoas and retroperitoneal abscesses |
| 11 | 53 year old female presenting with a 3-day history of difficulty breathing, face and neck swelling misdiagnosed with anaphylaxis. The physicians failed to consider that her symptoms with a history of rectal cancer and vascular port-a-cath situ, SVC obstruction was a likely diagnosis. | Anaphylaxis secondary to Henna | Superior vena cava syndrome |
| 12 | A 27-year-old presented with chest pain and a positive D dimer test. Was initially wrongly diagnosed with pulmonary embolism and started treatment. Subsequently, further testing showed that she actually had a pneumothorax, which the findings were originally missed in the chest XR. | Pulmonary embolism | Pneumothorax |
| 13 | A young woman was admitted for diabetic ketoacidosis with persistent, asymptomatic lactic acid elevation during the evolving COVID-19 pandemic. Cognitive biases in interpreting an elevated LA in this patient's care resulted in an extensive infectious workup instead of the low-cost and potentially diagnostic provision of empiric thiamine | DKA with infection | Thiamine deficiency |
| 14 | A 29-year-old lady who had a recent abortion and came in with abdominal pain. The initial diagnosis of endometritis for her subsequently was actually abdominal compartment syndrome and ischemic bowel. | Endometritis | Ischemic bowel |
| 15 | 6-year-old boy with myocarditis caused by invasive meningococcal infection, which was initially misdiagnosed as MIS-C due to diagnostic bias amid the COVID-19 pandemic as the patient | Acute myocarditis likely due to MIS-C | Acute myocarditis caused by invasive |

|    | fulfilled 5 of 7 of the CDC and 7 of 10 of the WHO criteria for MIS-C. |                           | bacterial infection |
|----|------------------------------------------------------------------------|---------------------------|---------------------|
| 16 | A 65-year-old lady presented to the ED for what seems like fluid overload which may be a result of congestive heart failure. Non-responsiveness to therapy was noted but physicians were unwilling to entertain differential diagnosis. Subsequently, the patient was found to have stage 4 ovarian serous adenocarcinoma. | Congestive cardiac failure | Malignancy |

RA; Rheumatoid Arthritis, TB; Tuberculosis, ED; Emergency Department, SIADH; Syndrome of inappropriate secretion of diuretic hormone, UTI; Urinary Tract Infection, HIV; Human Immunodeficiency Virus, SVC; Superior Vena Cava, MIS-C; Multisystem inflammatory syndrome in children, CDC; Centers for Disease Control and Prevention, WHO; World Health Organization.

**Multimedia Appendix 2.** PRISMA flow diagram for identification of case reports with cognitive bias.

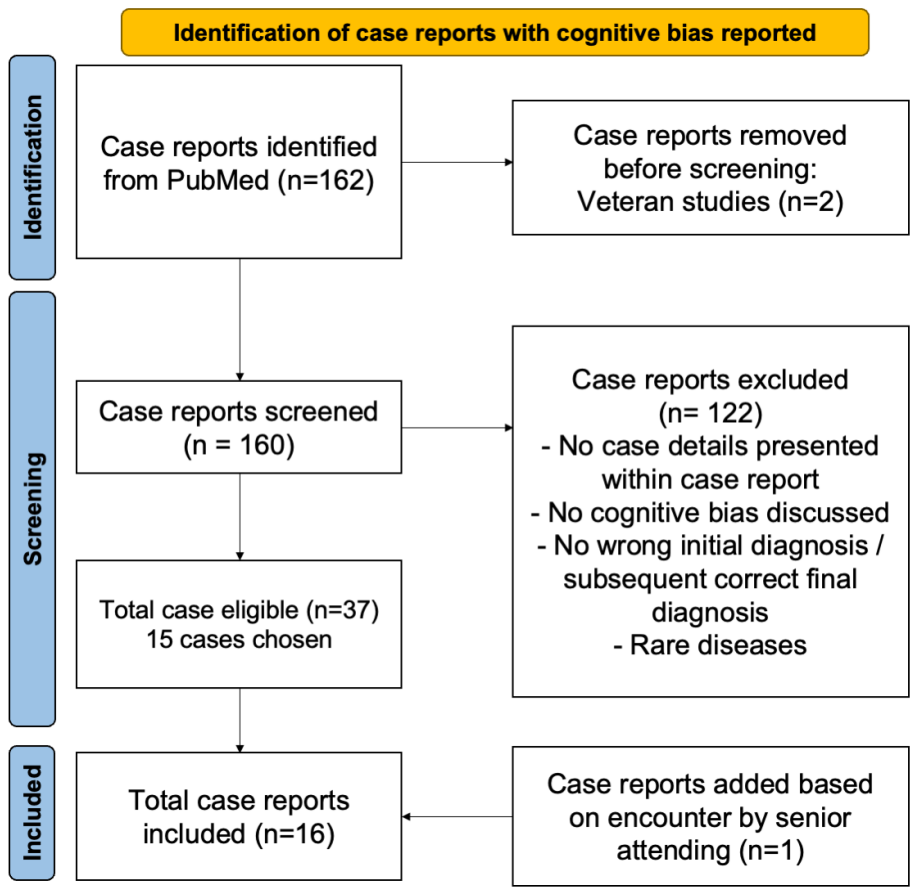